%% file: main.tex
\documentclass[10pt,twocolumn,letterpaper]{article}

\usepackage{cvpr}              % To produce the CAMERA-READY version
\input{preamble}
\definecolor{cvprblue}{rgb}{0.21,0.49,0.74}
\usepackage[pagebackref,breaklinks,colorlinks,allcolors=cvprblue]{hyperref}
\usepackage{xcolor}
\usepackage{listings}   
\usepackage{multirow}
\usepackage{algorithm}
\usepackage{algorithmic}
\usepackage{wrapfig}
\usepackage{hyperref}
\usepackage{url}
\usepackage{adjustbox}
\usepackage{graphicx}
\usepackage{amssymb}
\usepackage{subcaption}
\usepackage{array}      
\usepackage{booktabs}   
\usepackage{caption}   
\usepackage{float}
\usepackage{dblfloatfix}
\usepackage{xcolor}
\usepackage{subcaption}
\usepackage{tabularx}
\usepackage{makecell}
\usepackage[table]{xcolor}
\usepackage{pifont}
\usepackage[table,dvipsnames,svgnames]{xcolor}

\usepackage[pagebackref,breaklinks,colorlinks,allcolors=cvprblue]{hyperref}
\usepackage{xcolor}
\usepackage{listings}   
\usepackage{multirow}
\usepackage{algorithm}
\usepackage{algorithmic}
\usepackage{wrapfig}
\usepackage{hyperref}
\usepackage{url}
\usepackage{adjustbox}
\usepackage{graphicx}
\usepackage{amssymb}
\usepackage{subcaption}
\usepackage{marvosym}

\usepackage{array}      
\usepackage{booktabs}   
\usepackage{caption}   
\usepackage{float}
\usepackage{dblfloatfix}
\usepackage{xcolor}
\usepackage{subcaption}
\usepackage{tabularx}
\usepackage{makecell}
\usepackage[table]{xcolor}
\usepackage{pifont}
\usepackage[table,dvipsnames,svgnames]{xcolor}

\def\paperID{} % *** Enter the Paper ID here
\def\confName{CVPR}
\def\confYear{2026}

\title{DeepSketcher: Internalizing Visual Manipulation for Multimodal Reasoning}

\author{Chi Zhang\textsuperscript{1}, Haibo Qiu\textsuperscript{2$\dagger$}, Qiming Zhang\textsuperscript{3}, 
Zhixiong Zeng\textsuperscript{2},
\\
Lin Ma\textsuperscript{2\Letter}, Jing Zhang\textsuperscript{1\Letter} \\
\textsuperscript{1}School of Computer Science, Wuhan University, China  \\
\textsuperscript{2}Meituan Inc; 
\textsuperscript{3}Independent Researcher, China \\ 
\textbf{Project: \url{https://github.com/MiliLab/DeepSketcher}}
}

\begin{document}
\maketitle

\renewcommand\thefootnote{}%
\footnotetext{\textsuperscript{$\dagger$}Project Leader. 
\textsuperscript{\Letter}Corresponding Author: forest.linma@gmail.com, jingzhang.cv@gmail.com%
\addtocounter{footnote}{-1}%
}

\input{sec/0_abstract}    
\input{sec/1_intro}

\input{sec/2_relatedw}

\input{sec/3_method}
\input{sec/4_exp}

\input{sec/5_conclusion}
\input{sec/6_ack}

\input{sec/X_suppl}
\clearpage
{
    \small
    \bibliographystyle{ieeenat_fullname}
    \bibliography{main}
}

\end{document}

%% file: sec/0_abstract.tex
\begin{abstract}
The ``thinking with images'' paradigm represents a pivotal shift in the reasoning of Vision Language Models (VLMs), moving from text-dominant chain-of-thought to image-interactive reasoning. By invoking visual tools or generating intermediate visual representations, VLMs can iteratively attend to fine-grained regions, enabling deeper image understanding and more faithful multimodal reasoning. As an emerging paradigm, however, it still leaves substantial room for exploration in data construction accuracy, structural design, and broader application scenarios, which offer rich opportunities for advancing multimodal reasoning.
To further advance this line of work, we present DeepSketcher, a comprehensive suite comprising both an image–text interleaved dataset and a self-contained model. The dataset contains 31k chain-of-thought (CoT) reasoning trajectories with diverse tool calls and resulting edited images, covering a wide range of data types and manipulation instructions with high annotation accuracy. Building on this resource, we design a model that performs interleaved image–text reasoning and natively generates ``visual thoughts'' by operating directly in the visual embedding space, rather than invoking external tools and repeatedly re-encoding generated images. This design enables tool-free and more flexible ``thinking with images''. Extensive experiments on multimodal reasoning benchmarks demonstrate strong performance, validating both the utility of the dataset and the effectiveness of the model design.
\end{abstract}

%% file: sec/1_intro.tex
\begin{figure*}[t]
    \vspace{-2mm}
    \centering
    \includegraphics[width=0.9\linewidth]{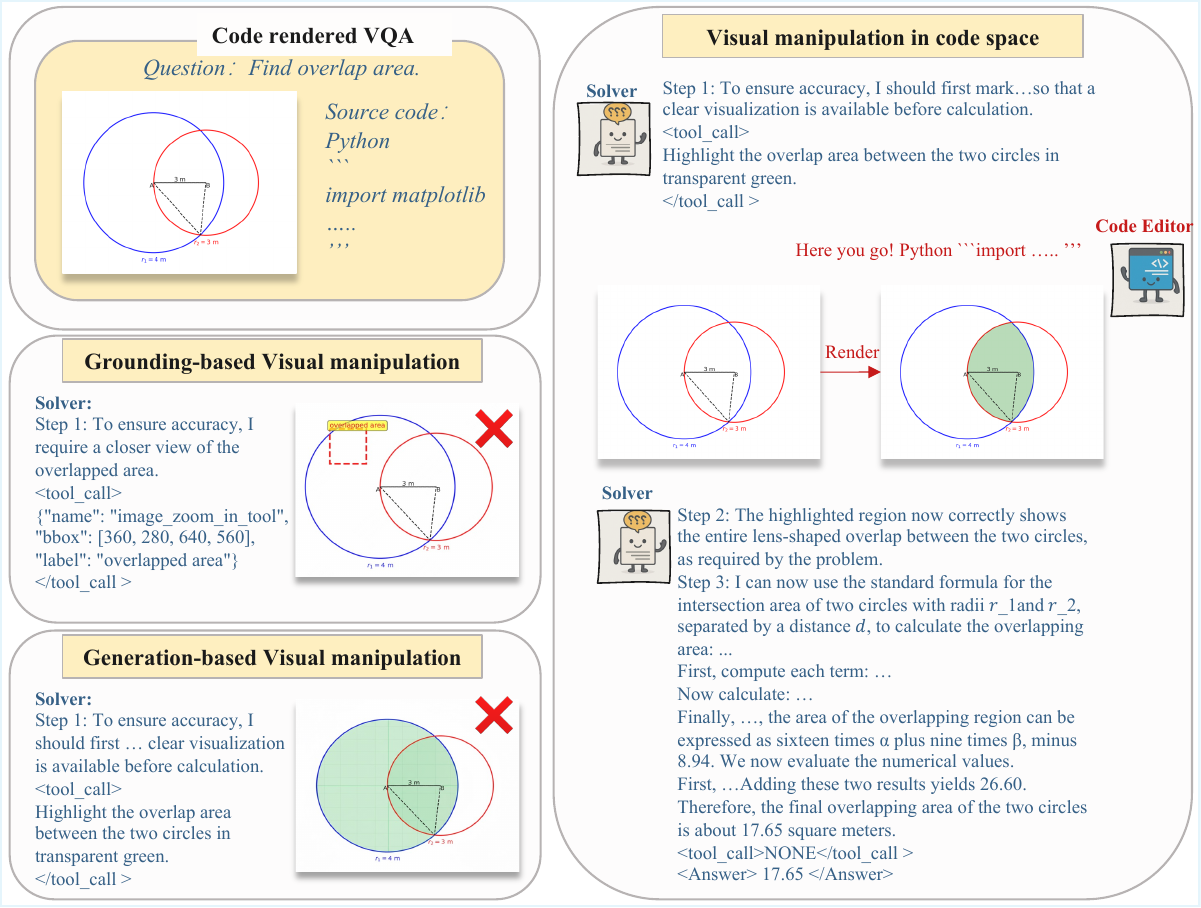}
    \caption{In code space (right), edits are specified through rendering code, offering precision and reproducibility. In contrast, grounding-based manipulation (bounding box predicted by GPT-5~\cite{openai2025gpt5}) and generation-based manipulation (image generated by Nano-Banana~\cite{nanobanana}) often yield noisy results, underscoring their limitations in stability and controllability. }
    \label{fig:diff}
    \vspace{-4mm}
\end{figure*}

\section{Introduction}
\label{sec:intro}
Recent progress shows that integrating step-by-step reasoning into VLMs has substantially improved their performance on complex tasks~\cite{meng2025mmeureka,yang2025r1onevision,coreteam2025mimovltechnicalreport,zhang2025r1,deng2025openvlthinker,chen2025sft,zhang2025perceptual,chen2026specs}. However, current VLMs often exhibit a ``thinking over seeing'' tendency~\cite{li2025self}: while they can generate lengthy and seemingly coherent reasoning traces, these traces are frequently detached from the actual visual input. In many cases, the models misinterpret critical details in the image or even hallucinate content that is not present~\cite{tu2025attention,sun2025mitigating}, suggesting that their reasoning is driven more by linguistic priors than by genuine visual perception~\cite{Guan_2024_CVPR,fu2025hidden}.

To address this, OpenAI introduced a new axis for VLM reasoning with ``thinking with images''~\cite{openai2025thinkingWithImages}. Instead of merely generating textual reasoning traces that overlook visual content, this approach enables models to actively interact with images through an explicit mechanism. By zooming, cropping, and performing image-level manipulations, VLMs are encouraged to ground their reasoning in actual visual evidence. This paradigm represents a shift from ``thinking over seeing'' to ``thinking through seeing,'' enabling models to analyze visual information more deeply, more thoroughly, and ultimately achieve reliable multimodal reasoning. Following such an idea, recent efforts have explored stimulating the use of visual information in the reasoning process to enhance model performance in perception and reasoning tasks. VILASR~\cite{wu2025reinforcing} defines a closed set of drawing operations and trains the model to decide when to invoke each of them. At inference time, the model selects an operation from this set and predicts the spatial coordinates required to execute it. DeepEyes~\cite{zheng2025deepeyes} and OPENTHINKIMG~\cite{su2025openthinkimg} leverage end-to-end reinforcement learning to incentivize ``thinking with images.'' In this setting, the model learns to actively manipulate visual inputs, such as zooming and cropping. Despite their differences, these approaches share a common limitation: the supported action space remains relatively restricted, and they inevitably rely on accurate spatial grounding, which remains challenging: curated data seldom yield perfectly accurate annotations, and end-to-end reinforcement learning rollouts are similarly error-prone. To overcome the constraints of a limited action space and to expand the model’s ``thinking space,'' another line of work makes a conceptual leap from execution to imagination, aiming to unify generation and reasoning within a single model~\cite{li2025zebra,li2025zebra,yang2025mirage}. However, this enlarged thinking space comes at the cost of extremely high training difficulty, and the methods' effectiveness has not yet been thoroughly validated on public benchmarks~\cite{qiao2024we,xiao2024logicvista,lu2023mathvista,wang2024measuring,zhang2024mathverse}. These methods provide promising directions for visual reasoning in VLMs, while also exposing fundamental trade-offs involving action space, grounding, training feasibility, and the inherent difficulty of constructing reliable data for supervision. 

\begin{figure*}[t]
\vspace{-1mm}
\centering
\begin{minipage}[t]{0.35\textwidth}
    \centering
    \includegraphics[width=0.8\textwidth]{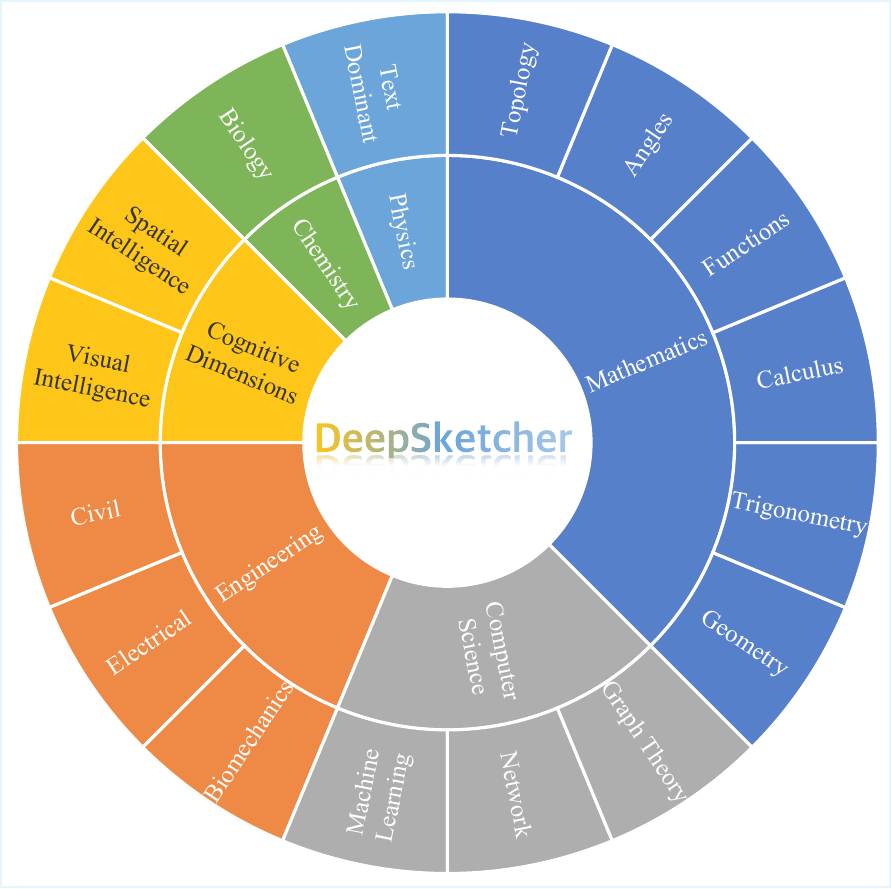}
    \captionof{figure}{Disciplinary coverage of our dataset.}
    \label{fig:cover}
\end{minipage}
\hfill
\begin{minipage}[t]{0.28\textwidth}
    \centering
    \includegraphics[width=\textwidth]{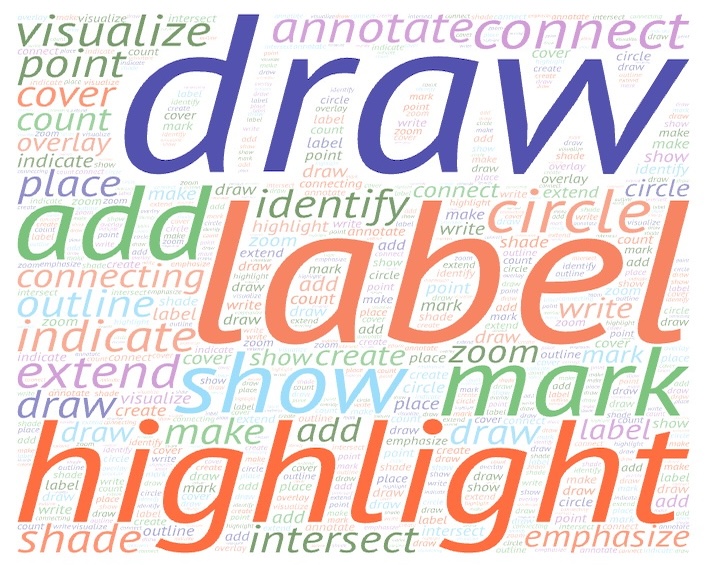}
    \captionof{figure}{Wordcloud of manipulations.}
    \label{fig:action}
\end{minipage}
\hfill
\raisebox{1.92cm}{%
\begin{minipage}[t]{0.35\textwidth}
    \centering
    \resizebox{\textwidth}{!}{
      \begin{tabular}{clrr}
        \toprule
        \textbf{Rank} & \textbf{Category} & \textbf{Count} & \textbf{Share (\%)} \\
        \midrule
        1 & Labeling/Annotation     & 12,340 & 20.9 \\
        2 & Highlighting            & 10,437 & 17.7 \\
        3 & Color Operations        & 7,383  & 12.5 \\
        4 & Circle Drawing          & 6,942  & 11.8 \\
        5 & Line Drawing            & 6,919  & 11.7 \\
        6 & Point Marking           & 3,924  &  6.6 \\
        7 & Area/Region Operations  & 2,641  &  4.5 \\
        8 & Shape Drawing           & 2,549  &  4.3 \\
        9 & Others                  & 5,919  &  4.3 \\
        \midrule
        \multicolumn{2}{r}{\textbf{Others}} & 4,853 & 10 \\
        \midrule
        \multicolumn{2}{r}{\textbf{Total}} & 59,054 & 100.0 \\
        \bottomrule
      \end{tabular}
    }
    \captionof{table}{Distribution of visual manipulations.}
    \label{tab:action}
\end{minipage}}
\vspace{-5mm}
\end{figure*}

To offer a complementary perspective in this paradigm, we introduce the DeepSketcher suite. The first component of the suite is a high-quality dataset with image–text interleaved chain-of-thought trajectories, where textual reasoning steps are interleaved with \texttt{<tool\_call>} instructions that return visually edited images, serving as auxiliary visual cues to guide subsequent reasoning. A distinctive feature of this dataset is that all images are code-rendered. Specifically, the source images are generated directly from rendering code, while intermediate images are obtained by modifying the source code according to the given instructions and re-rendering the updated code. This code-based approach provides both controllability and semantic clarity, enabling visual manipulations that are precise, reproducible, and less noisy than pixel-level editing, as illustrated by the running example in Figure~\ref{fig:diff}, which contrasts direct code-space editing with grounding-based and generation-based manipulations.
Beyond the complex ``reasoning $\rightarrow$ tool call instruction $\rightarrow$ image manipulation $\rightarrow$ reasoning'' pipeline in our dataset, we further propose a self-contained architecture to internalize the whole thinking mode into a single model.
Specifically, the model manipulates images directly in the visual embedding space, allowing seamless integration of visual and textual reasoning. This design removes the need for code execution, external tool calls, and repeated re-encoding of images, thereby enabling more flexible ``thinking with images'' patterns. In summary, the main contributions of this work are as follows:
\begin{itemize}
    \item By presenting the \textbf{DeepSketcher} suite, we provide a complementary perspective within the ``thinking with images'' paradigm, showing how dataset and model design can jointly support flexible multimodal reasoning.  
    \item We construct a high-quality dataset with interleaved image–text chain-of-thought trajectories. All images are code-rendered, and manipulations are conducted in code space, supporting a broad spectrum of open-vocabulary visual operations while avoiding the grounding noise inherent in previous datasets.  
    \item We design a self-contained model that internalizes the ``reasoning $\rightarrow$ tool call $\rightarrow$ image manipulation $\rightarrow$ reasoning'' chain. This design removes reliance on external tool calls, eliminates the need for coordinate-level predictions, and generalizes beyond code at inference.  
\end{itemize}

%% file: sec/2_relatedw.tex
\section{Related Work}
\label{sec:Related Work}
\subsection{Program-Based Frameworks for VQA}
Leveraging programmatic approaches to tackle complex visual problems is a rapidly emerging field with promising results~\cite{suris2023vipergpt,hu2024visual,lu2025octotools,zhang2025thyme,gupta2023visual}. 
Early work, such as ViperGPT~\cite{suris2023vipergpt}, solves VQA by prompting a code LLM to orchestrate external vision tools. Visual Sketchpad~\cite{hu2024visual} advances this by enabling VLMs to adapt their plans based on intermediate visual feedback, shifting the paradigm from static visual conditioning to dynamic reasoning. More recently, agentic frameworks like OctoTools~\cite{lu2025octotools} and REFOCUS~\cite{fu2025refocus} have introduced agentic systems that incorporate a broad suite of programmatic tools to boost proprietary model performance. While these frameworks provide a strong foundation for generating program-augmented trajectories, they typically rely on manipulating pixel-space via external tools. Separately, work like MathCoder-VL~\cite{wang2025mathcoder} has demonstrated the value of code's precision, using it to augment data and generate high-quality trajectories. However, this approach is primarily static. In contrast, DeepSketcher not only capitalizes on the rigorous nature of code-rendered data but also enables dynamic, interactive manipulation directly in the exact code space, eliminating grounding ambiguity and offering unparalleled precision.

\subsection{Vision Language Model Reasoning}
Enhancing the reasoning ability of vision-language models (VLMs) is a key focus of current VLM research. Following the success of GRPO~\citep{shao2024deepseekmath,guo2025deepseek} in textual reasoning, a growing body of work leverages reinforcement learning (RL) to strengthen the reasoning skills of VLMs, yielding promising progress~\citep{meng2025mmeureka,deng2025openvlthinker,yang2025r1onevision,zhang2025r1,coreteam2025mimovltechnicalreport,qiu2025metis,yang2025learning,zhong2025reading,lan2025metis,chen2026flexible,liu2026length}. Nevertheless, most existing approaches remain predominantly oriented toward textual reasoning steps, treating the visual input merely as a static condition. To move beyond simply seeing images and to reason more deeply about them, recent studies have introduced a tool-use paradigm, where external vision functions or specialized modules are invoked to manipulate visual inputs—for example, through cropping or zooming—and the edited artifacts are subsequently fed back into the model to guide the next stage of reasoning~\citep{zheng2025deepeyes,su2025openthinkimg}. This paradigm allows models to better perceive and localize fine-grained image regions, thereby improving VQA accuracy. Seeking to internalize these capabilities and eliminate the dependency on external tools, recent research explores equipping VLMs with native visual synthesis capabilities to actively shape their visual context~\citep{li2025zebra,yang2025mirage}. Yet, these efforts have largely been confined to limited scenarios such as jigsaw puzzles and mazes. DeepSketcher investigates this native generative paradigm within the multimodal mathematical domain.

%% file: sec/3_method.tex
\section{The DeepSketcher}
\subsection{The DeepSketcher Dataset}
\label{section: dataset}
\paragraph{Overview of the data curation pipeline.}
The DeepSketcher dataset contains extended interleaved image–text reasoning traces, where textual requests (e.g., highlighting a region or adding an auxiliary line) are followed by corresponding visual edits. Each trajectory establishes a loop between natural language reasoning and visual modification, encouraging models to ground their reasoning in visual evidence and fostering deeper multimodal understanding.

Prior approaches to constructing such data fall into two categories. (i) Grounding-based manipulation, where models predict an operation target, e.g., by outputting a structured action such as
$\{`name': `image\ zoom\ in\ ', `bbox': [360, 280, 640, 560]\}$ or by generating editing code to perform image modifications~\cite{hu2024visual,zheng2025deepeyes}. In both cases, the core mechanism relies heavily on precise coordinate regression. (ii) Generation-based manipulation, which leverages image generation models to fulfill editing instructions~\cite{chern2025thinking}. While promising, these paradigms struggle with grounding noise and limited precision, impeding the generation of consistent and controllable traces (Figure~\ref{fig:diff}).

In contrast, our approach builds upon code-rendered VQA data. Each instance is formulated as a tuple
$(C, I, Q, A),$
where $C$ represents the rendering code, $I = \mathcal{R}(C)$ is the image generated by the renderer $\mathcal{R}$, and $(Q,A)$ is the question-answer pair. This representation offers distinct advantages: (i) visual manipulations can be expressed as edits to $C$, ensuring accurate and reproducible modifications, (ii) the inherent code–image alignment eliminates spatial ambiguity and grounding noise, and (iii) the expressiveness of the code space supports a wide spectrum of operations, rather than limited to a predefined closed set. Based on this formulation, DeepSketcher dataset spans diverse domains (Figure~\ref{fig:cover}) and encompasses a wide variety of action-edit pair (Table~\ref{tab:action}). 

We employ a two-round pipeline to curate this dataset. In \emph{Round 1}, we design an automatic agentic system where two proprietary LLM experts collaborate to solve code-rendered VQA problems. Their interactions are compiled into 6k interleaved image–text CoT trajectories, serving as seed data to train an intermediate reasoning model capable of proper tool use and visual feedback interpretation. In \emph{Round 2}, we scale up data diversity by converting off-the-shelf VQA datasets into code-image pairs, expanding the dataset to 31k. The intermediate model is then deployed in the agentic system to generate richer trajectories  without incurring the prohibitive API cost of large-scale agentic interaction. These outputs train our final model (Section~\ref{section: model}). In the following, we detail the data collection pipeline and the agentic system.

\begin{algorithm}[t]
\caption{Agentic curation with \emph{Solver} (\,$\mathrm{LLM}_S$\,) and \emph{Code Editor} (\,$\mathrm{LLM}_E$\,)}
\label{alg:agentic}
\begin{algorithmic}[1]
\REQUIRE Initial code $C_0$, renderer $\mathcal{R}$, question $Q$, max steps $T_{\max}$
\STATE $I_0 \leftarrow \mathcal{R}(C_0)$; $\mathcal{D}_S \leftarrow \{Q\}$; $\mathcal{D}_E \leftarrow \emptyset$
\FOR{$t=0$ to $T_{\max}$}
  \STATE $(R_t, A, Act_t) \leftarrow \mathrm{LLM}_S(\mathcal{D}_S, I_t)$ \quad // CoT $R_t$; $A$ and $Act_t$ are mutually exclusive
  \STATE \textbf{assert} $(A=\varnothing)\ \oplus\ (Act_t=\varnothing)$ \quad // enforce exclusivity
  \STATE Append $(I_t, R _t, Act_t)$ to $\mathcal{D}_S$
  
  \IF{$A \neq \varnothing$}
    \STATE \textbf{return} $\mathcal{D}_S$ with $A$ \quad 
  \ELSE
    \STATE $C_{t+1} \leftarrow \mathrm{LLM}_E(C_{t}, Act_t, \mathcal{D}_E)$ \quad // complete edited code
    \STATE Validate $C_{t+1}$ (syntax/render checks); if invalid, repair or backoff
    \STATE $I_{t+1} \leftarrow \mathcal{R}(C_{t+1})$
    \STATE Append $(C_t, Act_t)$ to $\mathcal{D}_E$
  \ENDIF
\ENDFOR
\STATE \textbf{return} $\mathcal{D}_S$ with $A$ \quad 
\end{algorithmic}
\end{algorithm}

\paragraph{Data Collection.}
Our collection strategy adheres one fundamental principle: all images must be code-rendered. While such data may seem scarce, prior work has demonstrated its efficacy in enhancing VLM perception and reasoning across structured domains and beyond~\cite{jia2025chartreasoner,wang2025mathcoder,deitke2024molmo,cai2024geogpt4v,yang2025effective}. In the first round, we source data from  CoSyn-400k~\cite{yang2025scaling}, a large-scale dataset of code–image–QA triples. In the second round, we enhance diversity diversity by converting images from external datasets (e.g., MMK12~\cite{meng2025mmeureka}, UniGeo~\cite{chen2022unigeo}, MM-Math~\cite{sun2024mm}, GeoQA8k~\cite{chen2021geoqa}) into rendering code via an \texttt{img2code} pipeline. Rigorous verification and filtering steps ensure data validity (details in Appendix).

\begin{figure*}[!t]
    \vspace{-5mm}
    \centering
    \includegraphics[width=0.8\linewidth]{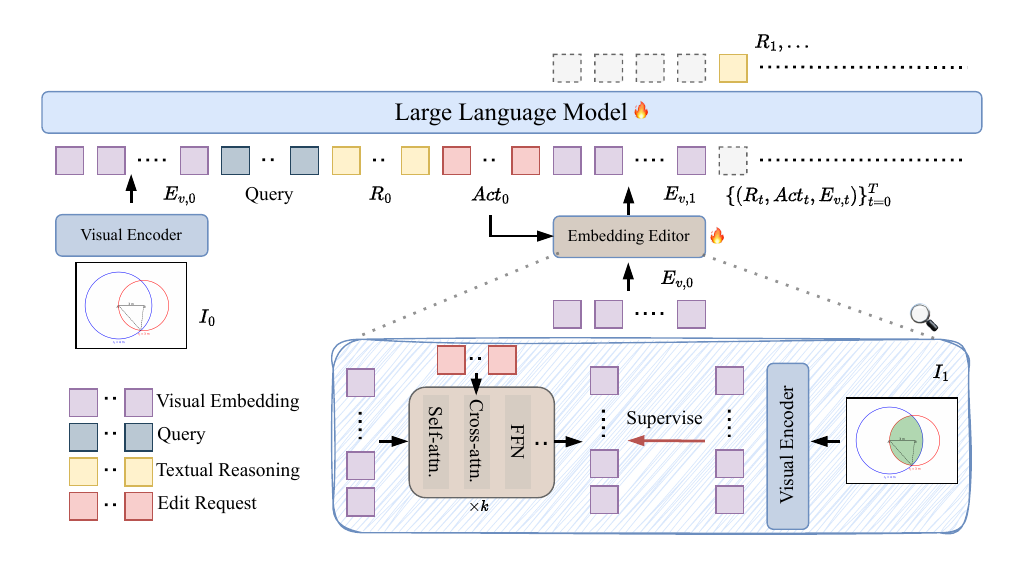}
    \vspace{-5mm}
    \caption{Architecture of the proposed DeepSketcher model. A query $Q$ and initial image $I_0$ are encoded into the vision–language model, producing reasoning tokens $R_t$ and edit instructions $Act_t$. The Embedding Editor manipulates visual embeddings directly, supervised by code-rendered ground-truth edits, and inserts updated embeddings back into the VLM context. This process yields interleaved reasoning and visual manipulation traces, ultimately producing the final answer.}
    \label{fig:model}
    \vspace{-3mm}
\end{figure*}

\paragraph{Agentic system for data curation.}
We curate reasoning traces using a two-agent collaborative framework. The system involves two LLM experts with complementary roles: a \emph{Solver} $\mathrm{LLM}_S$ that conducts step-by-step visual reasoning, and a \emph{Code Editor} $\mathrm{LLM}_E$ that edits rendering code according to natural-language instructions provided by $\mathrm{LLM}_S$. Specifically, given a code-rendered image $I_0 = \mathcal{R}(C_0)$ and a question $Q$, $\mathrm{LLM}_S$ is prompted to reason explicitly and, whenever visual evidence is uncertain or additional views are needed, to issue a free-form edit request $Act_t$ to $\mathrm{LLM}_E$ (e.g., ``draw a tangent line'', ``highlight point A in red''). Upon receiving the $Act_t$ and the current source code $C_{t}$, $\mathrm{LLM}_E$ returns a complete edited program $C_{t+1}$, which is rendered into a new image $I_{t+1} = \mathcal{R}(C_{t+1})$ and fed back to $\mathrm{LLM}_S$ for the next round of reasoning. The process continues until the \emph{Solver} produces a final answer $A$ or a termination condition is met (e.g., maximum edit steps). Then, we log the entire interleaved trajectory $\{(I_t, R_t, Act_t)\}_{t=0}^{T}$, where $R_t$ denotes the \emph{Solver}'s chain-of-thought at step $t$. This yields long image-text CoT traces aligned with code edits and rendered images, later standardized into training examples. To improve the reliability of the system, we incorporate mistake-proofing and verification mechanisms; full details are provided in Appendix. The overall procedure is summarized in Algorithm~\ref{alg:agentic}.

\subsection{The Deepsketcher Model}

\label{section: model}
\paragraph{Overview.}
The DeepSketcher dataset provides long-horizon, interleaved reasoning traces aligned with precise visual manipulations. To exploit this resource, we introduce the DeepSketcher model, an architecture designed to unify such interleaved reasoning with visual operations.

For comparison, common VLMs typically map an image $I$ and a textual query $Q$ to a sequence of textual reasoning steps $\{R_1, R_2, \ldots, R_t\}$ and a final answer $A$:
\[
\{R_1, R_2, \ldots, R_t\}, \; A = \mathrm{LLM}(E_v, Q),
\]
where $E_v$ is the embedding from the visual encoder.

In contrast, the DeepSketcher model treats reasoning and visual manipulation into a unified, dynamic trajectory ( Figure~\ref{fig:model}). Given an initial image-query pair $(I_0, Q)$, the visual encoder produces Embeddings $(E_{v,0}, Q)$. The model then generates an initial reasoning step $R_0$. When visual clarification is needed, it autonomously generates an action $Act_0$. The pair $(E_{v,0}, Act_0)$ is then processed by our built-in \emph{embedding editor} (Figure~\ref{fig:model}, bottom), which  predicts the manipulation directly in the visual embedding space, yielding an ``manipulated'' image representation $E_{v,1}$. The augmented context $\{E_{v,0}, Q, R_0, Act_0, E_{v,1}\}$ is then fed back into the model for subsequent reasoning. This recursive process produces an interleaved trajectory of reasoning, actions, and evolved visual states:
\begin{equation}
\begin{aligned}
&(R_0, Act_0, E_{v,1}, R_1, Act_1, E_{v,2}, \dots, \\
&\quad R_{T-1}, Act_{T-1}, E_{v,T}, A)
= \mathrm{DeepSketcher}(E_{v,0}, Q).
\end{aligned}
\end{equation}

Next, we detail the three-phase training strategy that enables this capability.

\paragraph{Building the DeepSketcher model.}

The training of the DeepSketcher model consists of three phases designed to progressively decouple the model from ground-truth visual inputs and enable autonomous latent editing.

\textbf{Phase 1: Reasoning Warm-up.} We initially optimize the model on interleaved image–text sequences using ground-truth image features, bypassing the embedding editor. The supervision signal is applied only to textual tokens and \texttt{<vision\_start>}, \texttt{<vision\_end>} tokens. These two special tokens serve as boundary markers for visual content in interleaved sequences. Visual features are inserted as continuous embeddings and serve only as conditioning context. This enables the model to learn proper structural demarcation between textual and visual modalities during generation. Formally, for the \(i\)-th training example with \(T^{(i)}\) images \(\{I^{(i)}_0,\dots,I^{(i)}_{T^{(i)}-1}\}\) interleaved with text. Each image \(I^{(i)}_t\) is encoded as a sequence of visual tokens \(E^{(i)}_{v,t}\), and we group the intervening text into segments \(\{\mathcal{S}^{(i)}_t\}_{t=0}^{T^{(i)}-1}\), where \(\mathcal{S}^{(i)}_t\) collects the positions of text tokens that appear after \(I^{(i)}_t\) and before \(I^{(i)}_{t+1}\). Let \(E^{(i)}_{v,\le t}=\{E^{(i)}_{v,0},\dots,E^{(i)}_{v,t}\}\) denote all visual tokens up to image \(t\). Each segment is modeled autoregressively, conditioning on the historical text \(x^{(i)}_{<\tau}\) and the preceding visual tokens \(E^{(i)}_{v,\le t}\), yielding the per-segment loss \(\mathcal{L}^{(i)}_t=-\sum_{\tau\in\mathcal{S}^{(i)}_t}\log P_\theta\!\big(x^{(i)}_\tau \mid x^{(i)}_{<\tau},E^{(i)}_{v,\le t}\big)\). The phase-1 language modeling objective then averages over all text tokens across the corpus and sums over examples, segments, and token positions:
\begin{equation}
\begin{split}
\mathcal{L}^{\text{phase-1}}_{\mathrm{LM}}(\theta)
&= - \frac{1}{\sum_{i=1}^N |\mathcal{S}^{(i)}|}
\sum_{i=1}^N \sum_{t=0}^{T^{(i)}-1} \sum_{\tau \in \mathcal{S}^{(i)}_t} \\
&\quad \log P_\theta\!\left(
x^{(i)}_\tau \,\middle|\, x^{(i)}_{<\tau},\, E^{(i)}_{v,\le t}
\right).
\end{split}
\label{eq:lm_phase1_triple}
\end{equation}

where \(|\mathcal{S}^{(i)}|=\big|\bigcup_{t=0}^{T^{(i)}-1}\mathcal{S}^{(i)}_t\big|\) is the number of text tokens in example \(i\). This objective trains the model to issue proper edit requests while ensuring that textual predictions are consistently conditioned on the available visual context.

\textbf{Phase 2: Embedding Editor Training.} The editor must handle a broad spectrum of visual modalities (e.g., geometry, charts) and follow diverse instructions to achieve reliable features. Thus, in the second phase, we suggest that larger scale and diverse supervision are indispensable to equip the model with native visual manipulation capabilities. Accordingly, we augment training data constructed via an \texttt{img2code} pipeline (detailed in Appendix) to capture the complexity of multimodal reasoning tasks, and we deploy the pretrained model for the agentic system described in Section~\ref{section: model}, which yields more training traces enriched with edit-request and image outcome pairs $(Act_t, I_{t+1})$. With these augmentations in place, we finalize the architecture to unify textual reasoning and visual manipulation.

Architecturally, when the model is uncertain about its visual perception, it generates an instruction enclosed by \texttt{<tool\_call>} tokens. We then extract the hidden states of these tokens, denoted $E_{\text{raw}} \in \mathbb{R}^{N\times D}$, and apply adaptive pooling to obtain a fixed-length sequence $E_{\text{act}} \in \mathbb{R}^{32\times D}$; the choice of 32 is chosen based on empirical statistics from training data. For the embedding editor, we adopt a Q-Former–style architecture~\cite{li2023blip} but drop the text branch and retain an image transformer with cross-attention. Unlike Q-Former, which grounds visual information into a fixed set of learnable query tokens, our module uses visual tokens themselves as queries and injects textual guidance from the action embeddings via cross-attention. Let $E_V \in \mathbb{R}^{K\times D}$ be the visual tokens from the frozen visual encoder. We take queries from $E_V$ and keys/values from $E_{\text{act}}$:
\[
Q = E_V W_Q,\qquad K = E_{\text{act}} W_K,\qquad V = E_{\text{act}} W_V,
\]
and update the visual tokens via a cross-attention block followed by an FFN:
\[
\widetilde{E}_V = \mathrm{MHA}(Q,K,V) + E_V,\qquad
E_V^{\mathrm{out}} = \mathrm{FFN}(\widetilde{E}_V) + \widetilde{E}_V.
\]
This design propagates instruction semantics directly into the visual space, yielding updated visual embeddings $E_V^{\mathrm{pred}} \in \mathbb{R}^{K\times D}$ with the same length $K$ as the input $E_V$.

We perform a second round of training on our proposed model. We initialize from the checkpoint of the reasoning model pretrained in the first stage and freeze all modules except the embedding editor. For supervision, we use the output of the visual encoder on ground-truth edited images as targets, and apply an $\ell_1$ loss to the latent editor’s predicted embeddings. Crucially, beginning in this phase, the VLM consumes editor-produced visual tokens rather than ground-truth visual context, and the LM objective is conditioned on the editor’s outputs. The phase-2 objective is:
\begin{equation}
\mathcal{L}^{\text{phase-2}}(\theta)
= \big\|E_V^{\text{pred}} - E_V^{\text{gt}}\big\|_1
\;+\mathcal{L}^{\text{phase-2}}_{\text{LM}}(\theta),
\label{eq:phase2}
\end{equation}
where $\mathcal{L}^{\text{phase-2}}_{\mathrm{LM}}(\theta)$ is the same as $\mathcal{L}^{\text{phase-1}}_{\mathrm{LM}}(\theta)$ except the visual embeddings. Compared to the prior approach~\cite{yang2025mirage} that edits images in a highly compressed latent space, our method preserves richer semantic information: the editor operates directly on visual tokens with explicit conditioning on action 
embeddings. This design yields more interpretable guidance and better semantic alignment between textual requests and visual transformations.

\textbf{Phase 3: Joint Adaptation.} Finally, we unfreeze the LLM backbone while maintaining the Phase 2 objective. This encourages the reasoning model to adapt to its own editor-produced visual outputs, ensuring consistency between the generated edit requests and the resulting visual context. The visual encoder is frozen throughout all stages.

\begin{table*}[t]
\vspace{-4mm}
\centering
\caption{Performance comparison on multimodel reasoning benchmarks.}
\resizebox{0.78\linewidth}{!}{
\begin{tabular}{lcccccc}
\hline
\textbf{Model} & \textbf{MathVerse} & \textbf{Mathvision} & \textbf{MathVista} & \textbf{LogicVista} & \textbf{WeMath} & \textbf{Average} \\
\hline
\rowcolor{gray!10}
\multicolumn{7}{c}{\small \textit{\textbf{Proprietary VLMs}}} \\
Claude3.7-Sonnet\ \cite{Anthropic2025Claude37} & 46.7 & 41.9 & 66.8 & 58.2 & 49.3 & 52.6 \\
GPT-4.1 \ \cite{openai2024gpt41}          & 48.9 & 46.4 & 70.4 & 61.1 & 55.5 & 56.5 \\
\hline
\rowcolor{gray!10}
\multicolumn{7}{c}{\small \textit{\textbf{Open-source VLMs}}} \\
InternVL3-8B\ \cite{zhu2025internvl3}       & 38.5 & 26.3 & 70.4 & 45.6 & 31.7 & 42.5 \\
Qwen2.5-VL-7B\ \cite{bai2025qwen2}    & 41.1 & 27.0 & 68.2 & 39.8 & 34.3 & 42.1 \\
Mulberry-Qwen-7B\ \cite{yao2024mulberry} & 35.9 & 27.6 &65.1 & 40.7 & 28.7 & 39.6 \\
MathCoder-VL\ \cite{wang2025mathcoder} & 37.0 & 22.3 & 55.1 & 32.4 & 39.4 & 37.2\\
\hline
\rowcolor{gray!10}
\multicolumn{7}{c}{\small \textit{\textbf{Tool-Calling VLMs}}} \\
VILASR-7B\ \cite{wu2025reinforcing}         & 29.4  &  25.0 &  57.6 &  32.2 &  23.7 & 33.6  \\
DeepEyes-7B\ \cite{zheng2025deepeyes}       & 42.2 & 26.6 & 70.1 & 47.7 & 38.9 & 45.1 \\
Thyme-VL-7B\ \cite{zhang2025thyme} & 39.1 & 27.6 & 70.0 & 49.0 & 39.3 & 45.0 \\
\hline
\rowcolor{gray!10}
\multicolumn{7}{c}{\small \textit{\textbf{Inner Visual Thought VLMs}}} \\
Bagel-Zebra-CoT-7B\ \cite{li2025zebra}  & 48.8  &  28.2 &  64.7 & 48.4  &  28.0 & 43.6  \\
Mirage-7B\ \cite{yang2025mirage}          &  27.3 & 28.6  &  63.7 &  40.7 &  16.7 &  35.4 \\
DeepSketcher-7B (Ours)           & 43.2 & 32.3 & 69.1 & 48.1 & 37.1 & 46.0 \\
\textcolor{red}{$\Delta$ (vs Qwen2.5-VL-7B)}            & \textcolor{red}{\textit{+2.1}} & \textcolor{red}{\textit{+5.3}}& \textcolor{red}{\textit{+0.9}}& \textcolor{red}{\textit{+8.3}}& \textcolor{red}{\textit{+2.8}}&\textcolor{red}{\textit{+3.9}}\\
\hline
\end{tabular}}
\label{tab:math_models}
\vspace{-3mm}
\end{table*}

%% file: sec/4_exp.tex
\section{Experiments}
\subsection{Setups}
\label{sec: expsetups}
\paragraph{Baselines.}
To evaluate the effectiveness of the proposed DeepSketcher model, we compare it against four categories of baselines: (1) proprietary models, including Claude3.7-Sonnet~\citep{Anthropic2025Claude37} and GPT-4.1~\citep{openai2024gpt41}; (2) state-of-the-art open-source models~\citep{zhu2025internvl3, bai2025qwen2} and reasoning VLMs~\cite{yao2024mulberry,wang2025mathcoder}; (3) reasoning VLMs with tool-calling capabilities that rely on external tools~\citep{zheng2025deepeyes,wu2025reinforcing}; and (4) ``thinking-with-generated-images'' models that produce inner visual thoughts~\citep{li2025zebra,yang2025mirage,zhang2025thyme}. Strictly speaking, our model also falls into the fourth category, as it performs interleaved visual-textual reasoning natively, without external tools. We select Qwen2.5-VL-7B as our baseline model.

\paragraph{Benchmarks.} 
We evaluate our model on common multimodal reasoning benchmarks, including MathVerse (vision-only)~\citep{zhang2024mathverse}, MathVision (mini)~\citep{wang2024measuring}, MathVista (mini)~\citep{lu2023mathvista}, LogicVista (Overall)~\citep{xiao2024logicvista}, and WeMath (Overall)~\citep{qiao2024we}. We also construct an in-house benchmark, Indicator-500, by sampling 500 code-rendered VQA instances from the Cosyn-400k test set (which also go through our data verification and filtering process). Unlike existing benchmarks, it includes paired code information, which enables to decouple interleaved reasoning from visual manipulation and provides a reliable indicator {for the embedding editor during training. (See the Appendix for training details.)}

\subsection{Main Results}
Table~\ref{tab:math_models} summarizes the performance of different VLMs across the benchmarks described above. For clarity, we group the baselines into four categories as mentioned earlier. The last group, ``Inner Visual Thought VLMs,'' is particularly challenging, as its ``thinking space'' and ``action space'' are far larger than those of tool-calling VLMs with fixed utilities. Such models are more sensitive to visual variations, and their robustness in visual manipulation may affect the model performance. When they fail to generate reliable visual content, the resulting noise can propagate through the reasoning trace and hurt overall performance. Despite these challenges, our model consistently outperforms other inner visual thought VLMs such as Bagel-Zebra-CoT-7B and Mirage-7B across most benchmarks. Furthermore, when compared with tool-calling VLMs, despite operating under a substantially more flexible paradigm, it surpasses VILASR-7B and DeepEyes-7B by 12.4 and 0.9 points in average. Together, these results highlight the effectiveness of our approach within this challenging setting, which we attribute to the accuracy and reliability of our training data, and the adaptability of our proposed model architecture.

To better understand the effect of our method, we conduct an in-depth comparison against the base model Qwen2.5-VL-7B. Overall, our approach yields an average improvement of 3.9 points across benchmarks. When breaking down results by task category, consistent patterns emerge: the most reliable gains appear in tasks involving geometry and counting, with particularly striking improvement on MathVision reaching 5.3 points. In addition, math-related problems (LogicVista) involving logical or numerical reasoning also exhibit significant improvements (8.3 points). By contrast, the improvement on tasks that require symbolic manipulation or domain knowledge integration tend to decline, decreasing performance gain to 0.9 points on MathVista. In particular, this dataset contains scientific reasoning and textbook QA, both of which have numerous open-domain images and depend heavily on disciplinary knowledge outside the scope of our training data. Please see the Appendix for more performance details.

\begin{table*}[t]
\centering
\vspace{-3mm}

% ---------- 左表 ----------
\begin{minipage}[t]{0.4\textwidth}
\centering
\caption{Comparison of collaborative vs. independent answers across different LLMs.}
\resizebox{0.8\linewidth}{!}{
\begin{tabular}{llc}
\toprule
\emph{Solver}  & \emph{Code Editor}  & \emph{pass@8} \\
\midrule
GPT-4.1       & Null   & 0.72 \\
GPT-4.1       & Claude3.7-Sonnet    & 0.80 \\
Qwen2.5-VL-72B   & Null   & 0.67 \\
Qwen2.5-VL-72B   & Claude3.7-Sonnet    & 0.72 \\
\bottomrule
\end{tabular}}

\label{tab:agentic}
\end{minipage}
\hfill
% ---------- 右表 ----------
\begin{minipage}[t]{0.56\textwidth}
\centering
\vspace{3.7mm}
\caption{Ablation study on the embedding editor.}
\resizebox{\linewidth}{!}{
\begin{tabular}{llccccccc}
\toprule
\textbf{Stage} & \textbf{Setting} & \textbf{Mathverse} & \textbf{Wemath} & \textbf{Mathvista} & \textbf{Mathvision} & \textbf{LogicVista} & \textbf{Indicator-500} & \textbf{Average} \\
\midrule
\multirow{3}{*}{2} 
 & Text-only (\emph{Baseline}) & 37.2 & 28.3 & 65.0 & 28.6 & 45.9 & 38.3 &40.6\\
 & Editor   & 41.6 & 37.5 & 65.8 & 28.9 & 46.5 & 33.8 & 42.4\\
 & Agentic (Oracle) & -- & -- & -- & -- & -- & 41.0 & -- \\
\midrule
\multirow{3}{*}{3} 
 & Text-only (\emph{Baseline}) & 38.1 & 31.2 & 65.7 & 33.5 & 41.8 & 37.5 & 41.3\\
 & Editor  & 43.2 & 37.1 & 69.1 & 32.3 & 48.1 & 40.5 & 45.1\\
 & Agentic (\emph{Oracle}) & -- & -- & -- & -- & -- & 44.8 & -- \\
\bottomrule
\end{tabular}}

\label{tab:model_comparison}
\end{minipage}

\end{table*}

\begin{figure*}[t]
    \centering
    \includegraphics[width=1\linewidth]{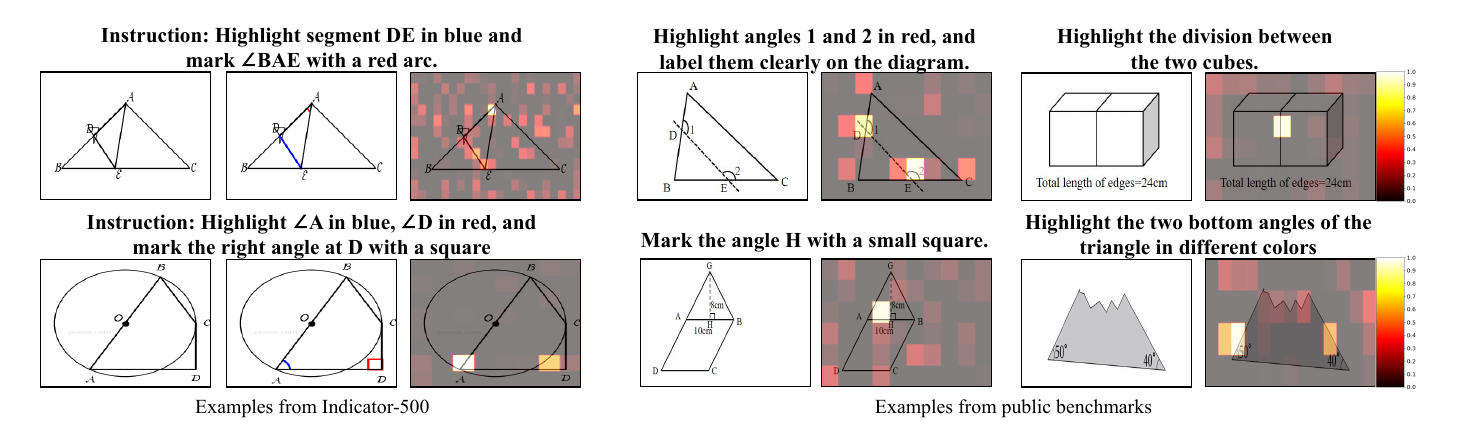}
    \vspace{-6mm}
    \caption{Difference map visualizations. Each example shows the input image (left), the programmatic rendering (available only in Indicator-500) (middle), and the difference map between the embedding editor output and the original visual embedding (right).}
    \label{fig:roi}
    \vspace{-3mm}
\end{figure*}

\subsection{Ablation Study}
\label{sec: ablate}
\paragraph{Ablation study on the agentic data curation system.}
For data curation, we design an agentic system where two experts collaborate to solve VQA. The intuition is that, with the aid of a \emph{Code Editor} that has direct access to the source code underlying an image, the solver LLM effectively gains more accurate visual information. With this enhanced context, the solver can tackle problems that are otherwise unsolvable via independent reasoning, thereby enabling the collection of higher-quality and more informative interleaved reasoning traces. To verify this, we conduct a controlled experiment on a subset of Cosyn-400k under two settings: 
(i) an \emph{independent answering} setup, where solver LLM works alone, and (ii) a \emph{collaborative answering} setup, where the solver cooperates with an LLM acting as the \emph{Code Editor}. We evaluate the results using the \emph{pass@8} metric, which counts a question as correctly answered if at least one of the eight responses matches the ground truth. As shown in Table~\ref{tab:agentic}, both solvers achieve significant gains when paired with the \emph{Code Editor}, confirming that collaboration enables the agentic system to resolve more challenging questions beyond the reach of single models. Consequently, our pipeline retains more verified examples increasing coverage and capturing reasoning traces from complex problems that would otherwise be discarded. These trajectories are not only richer in content but also offer greater value for training.

\paragraph{Does the model reason better with embedding editor?}
\label{embed editor}
To evaluate the efficacy of the embedding editor, we conduct an ablation study across multiple multimodal reasoning benchmarks and our in-house Indicator-500 evaluation set (Table~\ref{tab:model_comparison}). We focus on the models from training stage 2 and 3, as the stage 1 model lacks the embedding editor. For each model, we compare three experimental settings: (i) deploying the model within an \emph{agentic system} that collaborates with an external \emph{Code Editor} expert, which has direct access to the source code of the input image and thus makes all information explicitly available, this approximates an \emph{upper bound}; (ii) a setup relying solely on the model’s built-in embedding editor to manipulate visual representations and generate interleaved reasoning traces; and (iii) a text-only \emph{baseline} that  bypasses the editor (i.e., the model receives the original embeddings regardless of its edit requests) to produce chain-of-thought traces. Any degradation relative to this baseline would imply that the editor introduces noise rather than enhancing reasoning.

As shown in Table~\ref{tab:model_comparison}, the \emph{embedding editor} consistently outperforms the text-only baseline on most benchmarks. A notable exception occurs in stage 2, where the editor-equipped model trails the baseline by 4.5 points on Indicator-500. This issue is largely alleviated in stage 3, where joint training of the LLM and the editor tightens their coupling. In stage 3, we observe gains under all settings: the text-only baseline improves slightly (40.6 to 41.3), adding the editor lifts the average further to 45.1 (+3.8), and the agentic upper bound reaches 44.8 on \emph{Indicator-500}. Overall, later-stage joint training reduces the distribution shift from editor-produced visual tokens, yielding broad performance gains, and improving visual grounding reliability. The remaining gap to the agentic upper bound underscores opportunities for further architectural optimization.

We also characterize how the embedding editor modifies visual content. Since the internal visual outputs are not directly observable, we measure the distance between the edited outputs and the input image embeddings, localizing pronounced differences as regions of interest (ROIs). We perform this analysis on both in-house Indicator-500 dataset and public benchmarks. 
For Indicator-500 (Figure~\ref{fig:roi}, left), the availability of the underlying source code allows us to issue identical instructions to the \emph{Code Editor} expert to obtain programmatically rendered edits for comparison. We observe that the ROIs generally align well with both the natural language instructions and the programmatic edits. It is worth noting that while programmatic edits serve as a strong reference, they are  illustrative rather than absolute ground truth, as natural language instructions may admit multiple valid implementations. On the public benchmarks (Figure~\ref{fig:roi}, right), where no programmatic edits are available. The ROIs still largely align with the intent expressed in the natural language instructions. This reinforces the effectiveness of our proposed module and demonstrates its generalization capability beyond the code-accessible setting.

%% file: sec/5_conclusion.tex
\section{Conclusion}
We present DeepSketcher as a fresh perspective within the broader paradigm of ``thinking with images.'' At the heart of this suite lies a carefully constructed dataset, where chain-of-thought reasoning is interleaved with code-rendered visual edits—precise, reproducible, and semantically grounded. Building upon this foundation, we design a self-contained model that internalizes the entire cycle of reasoning, tool invocation, and image manipulation. By removing reliance on external tools and fragile coordinate predictions, the model demonstrates a new pathway toward a resilient multimodal intelligence. Together, these contributions point toward a future where machines learn to ``think'' with images in a more integrated way.

%% file: sec/6_ack.tex
\section{Acknowledgment}
This work was supported by the New Generation Artificial Intelligence-National Science and Technology Major Project (No. 2025ZD0123602). The numerical calculations in this paper have been done in part on the supercomputing system in the Supercomputing Center of Wuhan University.

%% file: sec/X_suppl.tex
\clearpage
\setcounter{page}{1}
\maketitlesupplementary

\noindent\textbf{Overview.} This supplementary material begins by providing a comprehensive review of related work (Sec.~\ref{sec:app related work}) and elaborating on the details of dataset construction (Sec.~\ref{sec:data_details}) and training configurations (Sec.~\ref{sec:training_details}). We then present extensive discussions (Sec.~\ref{sec:discussions}), which include additional visualization results, ablation studies comparing interleaved versus pure-text CoT and decoupled training strategies, as well as a specific performance analysis on MathVerse. Finally, we discuss the limitations of our approach and outline directions for future work (Sec.~\ref{sec:limitations}).

\section{More Related Work}
\label{sec:app related work}
Here, we provide additional related work, supplementing Sec.~\ref{sec:Related Work}.
\subsection{Visual Prompting}
Visual Prompting (VP) is an established interaction paradigm predating the recent surge in multimodal reasoning. This approach involves incorporating pixel- or region-level cues—such as bounding boxes, markers, scribbles, or segmentation masks—into input images. Prior studies demonstrate that this technique substantially enhances a model’s perceptual capabilities and overall visual understanding~\citep{yang2023set,hu2024visual,cai2023vipllava,yan2024list}. 

For instance, SoM~\citep{yang2023set} showed that augmenting input images with labeled cues significantly improves the referring and localization performance of GPT-4V. Similarly, Sketchpad-style pipelines~\citep{hu2024visual} automatically compose visual prompts by leveraging a toolbox of detectors and segmenters (often orchestrated by lightweight scripts) to draw boxes and masks either pre-inference or during inference. This process strengthens both perception and downstream reasoning.

However, these VP methods were primarily developed for proprietary frontier models. This reliance stemmed from the fact that many open-source alternatives lacked the capacity to reliably invoke external tools or effectively reason over the resulting feedback. This limitation underscores the critical need for models to internalize these skills, rather than depending solely on externally orchestrated prompting mechanisms.

\begin{figure*}[!t]
    \centering
    \includegraphics[width=0.7\linewidth]{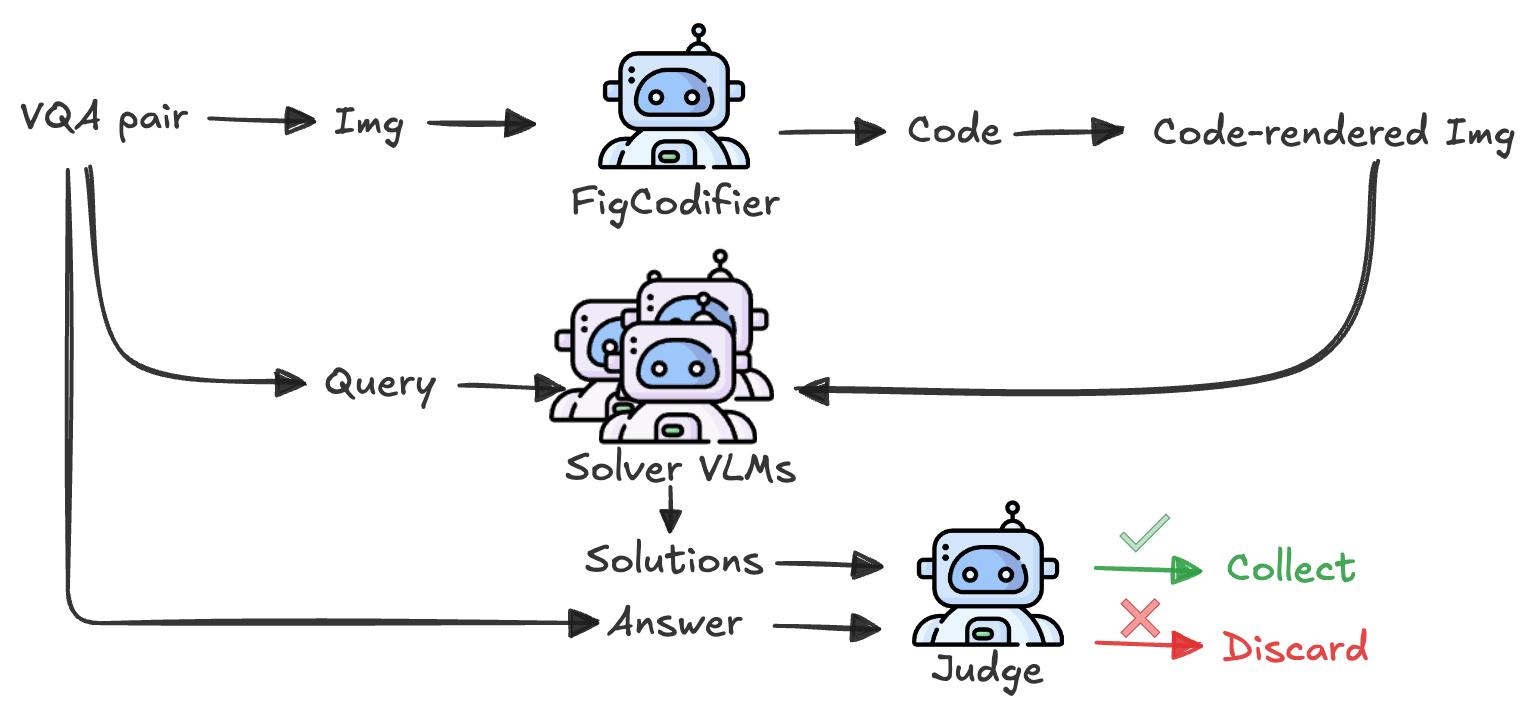}
    \caption{Overview of \texttt{img2code} pipeline.}
    \label{fig:Img2code}
\end{figure*}

\section{More Details on Dataset Construction}
\label{sec:data_details}
 Here, we detail the data preparation schemes employed to ensure our training trajectories are less noisy, higher in quality, more diverse, and better aligned with real-world, human-authored problems. This process includes our \texttt{img-to-code} pipeline, data filtering schemes, and the final prompt template.

\subsection{Img-to-code Pipeline}
\label{app:img2code}
Current code-rendered datasets suffer from an inherent drawback: the distribution of synthesized questions differs substantially from those authored by human experts in high-school, university, or competition settings. Moreover, it is difficult to reliably assess the difficulty of these generated problems. The perspectives adopted in question construction often lack the nuance and pedagogical intent typically found in human-authored questions. This gap is critical, as the core objective for advanced reasoning models is to solve challenging, real world problems.

To bridge this gap, we designed an \texttt{img2code} pipeline to incorporate more realistic and challenging problems during our data expansion round. An overview of this pipeline is shown in Fig.~\ref{fig:Img2code}. Concretely, we first sample VQA data from a collection of math-dominant datasets~\citep{meng2025mmeureka,chen2022unigeo,sun2024mm,chen2021geoqa}. We then employ FigCodifier~\citep{wang2025mathcoder}, a model specifically trained to convert images into code, to process these samples. The resulting code is subsequently rendered back into images for verification.

It is worth noting that \texttt{img2code} is an extremely challenging task, as it requires the model to faithfully capture all fine-grained details in an image using programmatic language. 
We conduct preliminary tests using three models: FigCodifier, GPT-4.1, and Claude 4.0-Sonnet~\citep{Anthropic2025Claude4}. By human inspection, the success rates of all three are below 10\%. This low success rate is attributable to the task's sensitivity; even a minor error in the rendering process can lead to a drastically different semantic meaning. Therefore, we employ a pragmatic compromise for automatic quality filtering. Specifically, we leverage multiple solver VLMs, including GPT-4.1, Claude 3.7-Sonnet, and Qwen2.5-VL-72B, to answer the VQA questions using the re-rendered images. If a solver model can produce a correct answer, we deem the re-rendered image acceptable. This indicates that while the image might not be a perfect replication, it retained sufficient semantic information for the problem-solving task.

\begin{figure*}[t]
    \centering
    \includegraphics[width=0.7\linewidth]{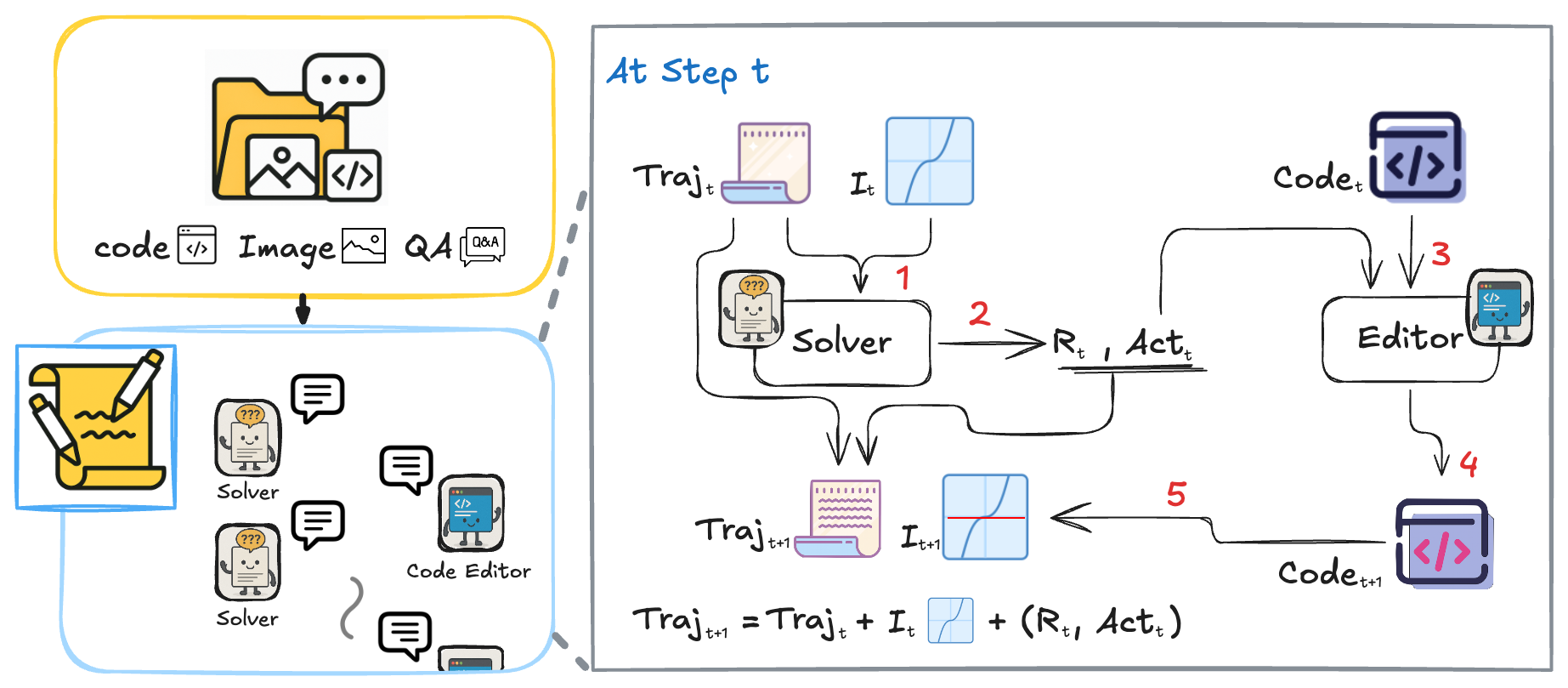}
    \caption{Overview of the DeepSketcher data curation pipeline. We first construct a dataset of VQA problems with images rendered directly from code. An agentic system is then designed to generate interleaved image–text reasoning traces.}
    \label{fig:agentic}
    \vspace{-4mm}
\end{figure*}

\subsection{Data Filtering}
Here, we describe our data filtering process in detail. In the first round of curation, we selected samples from the math, graphic, diagram, and chart subsets of the Cosyn-400k dataset. Although this dataset provides a large volume of diverse data, all samples are synthesized: both questions and answers are generated by LLMs. Although CoSyn has gone through rigorous data scrutiny, to further ensures the validity of the provided answers, we employ several LLM experts, including GPT-4.1, Qwen2.5-72B-VL, and GPT-4.1-mini, to independently answer each question. For every question, we sample two responses from each LLM. If at least one of these responses matches the original answer provided by Cosyn-400k, we retain the question–answer pair; otherwise, we discard it.  

For the agentic system, we design several fail-safe loops and verification strategies. First, if the code produced by the \emph{Code Editor} fails to execute during rendering, the erroneous code together with the error logs are sent back to the editor for another round of editing. By leveraging the error logs, the editor can dynamically adjust its edits, thereby mitigating issues arising from either model mistakes or inconsistencies in the execution environment. If the code is rendered successfully, we then prompt the \emph{Solver} LLM to critically inspect and challenge the rendered content rather than simply accepting it. If the content does not satisfy the \emph{Solver}'s requirements, the \emph{Solver} generates revised instructions for another round of editing. 

Finally, before model training, we apply a rejection sampling strategy~\citep{dong2023raftrewardrankedfinetuning}. Specifically, we use the base model Qwen2.5-VL-7B to answer all queries and discard those for which it produces the correct answer, retaining only the more challenging cases for training.
\subsection{Prompt Template}
\begin{figure*}
    \centering
    \includegraphics[width=1\linewidth]{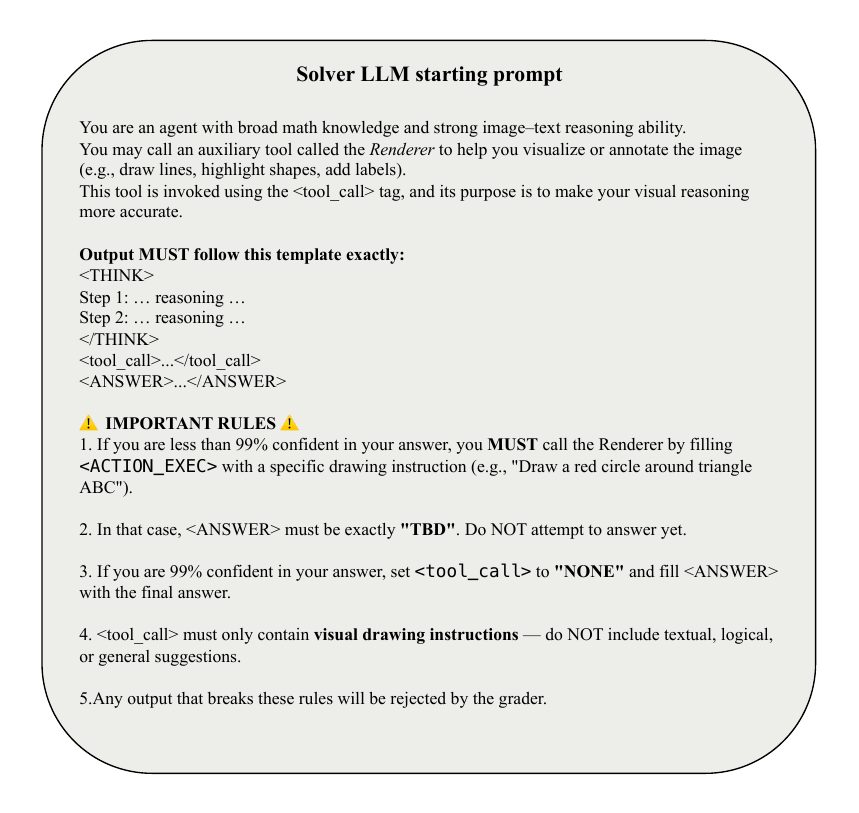}
    \caption{The \emph{Solver} LLM starting prompt.}
    \label{fig:prompt1}
\end{figure*}

\begin{figure*}
    \centering
    \includegraphics[width=1\linewidth]{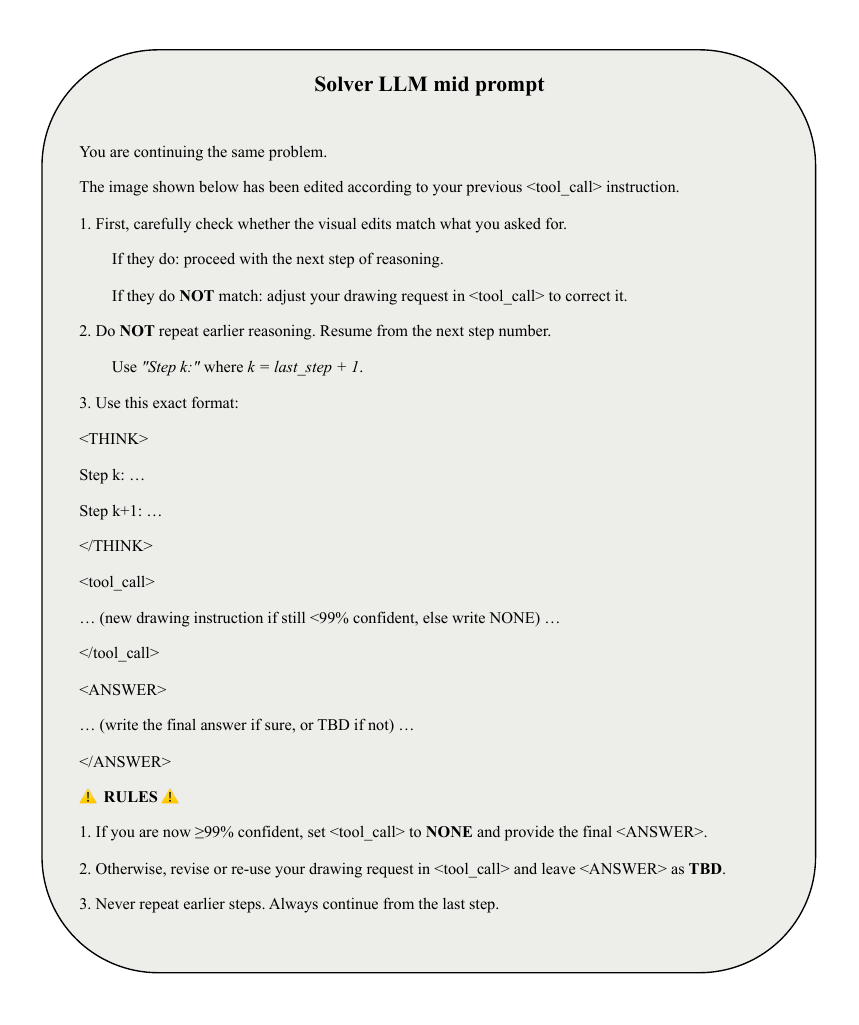}
    \caption{The prompt template when \emph{Solver} LLM receives updated visual information.}
    \label{fig:prompt2}
\end{figure*}

\begin{figure*}
    \centering
    \includegraphics[width=1\linewidth]{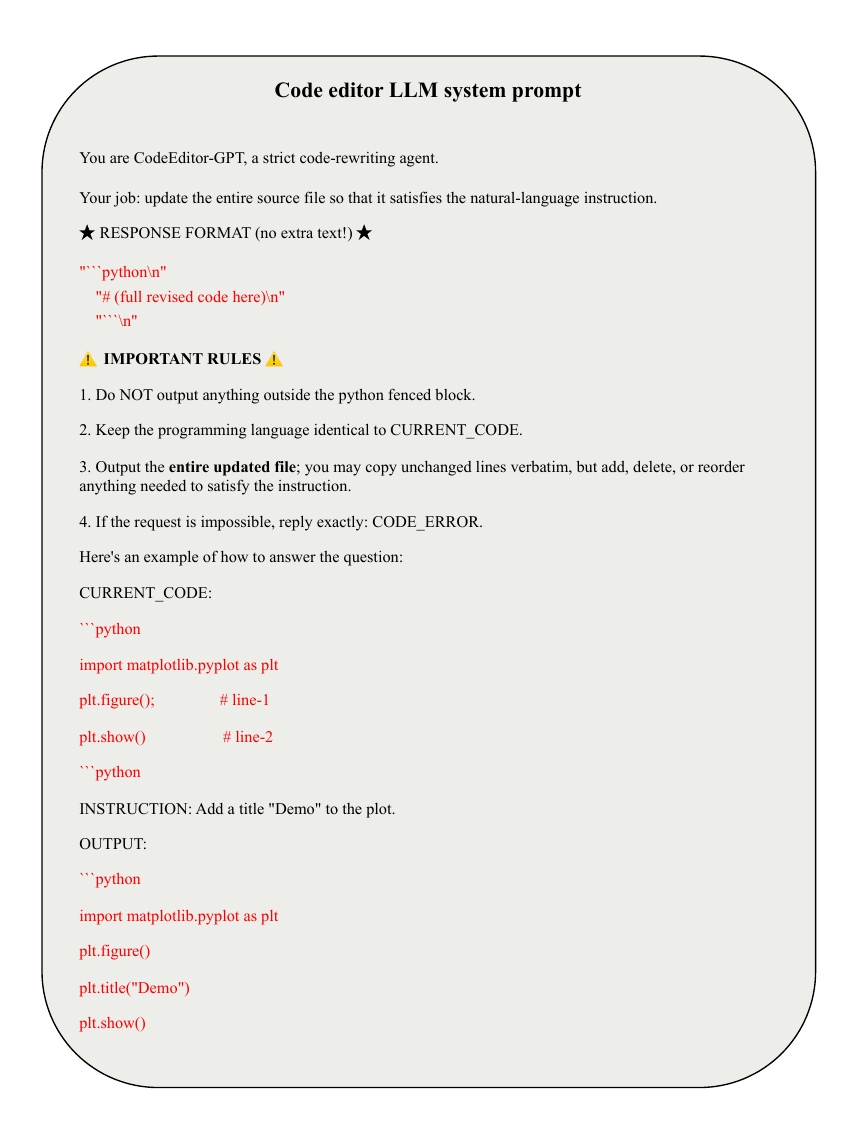}
    \caption{The prompt template for \emph{Code Editor} LLM.}
    \label{fig:prompt3}
\end{figure*}

We show the prompt template used in the agentic data curation system in Fig.~\ref{fig:prompt1}, Fig.~\ref{fig:prompt2}, and Fig.~\ref{fig:prompt3}.

\begin{figure}[!ht]
    \centering
    \begin{subfigure}[t]{\linewidth}
        \centering
        \includegraphics[width=1\linewidth]{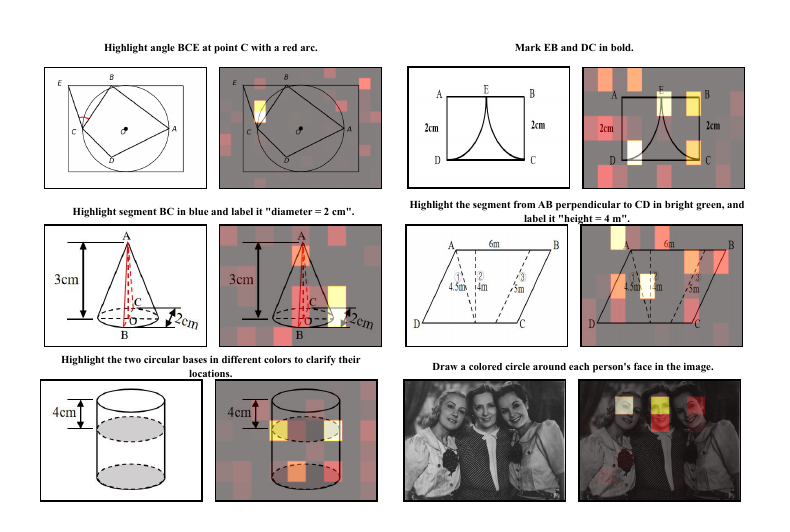}
        \caption{Cases where the model generally attends to the region in accordance with the instruction.}
        \label{fig:roipublic}
    \end{subfigure}
    \hfill
    \begin{subfigure}[t]{\linewidth}
        \centering
        \includegraphics[width=\linewidth]{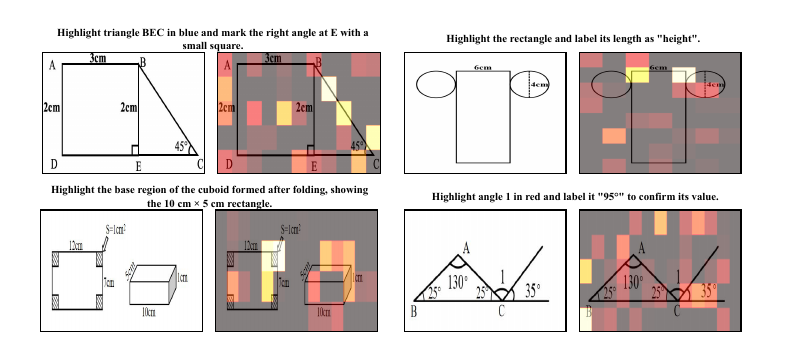}
        \caption{Cases where the model’s attention deviates from its textual intent.}
        \label{fig:failroi}
    \end{subfigure}
    \caption{Difference map visualization from public benchmarks. (a) Alignment between attention and instruction. (b) Cases with deviation from textual intent.}
    \label{fig:roi_public_fail}
\end{figure}

\section{Training Details}
\label{sec:training_details}
We adopt Qwen2.5-VL-7B~\citep{bai2025qwen2} as the base model. Our implementation is built on LLaMA-Factory~\citep{zheng2024llamafactory}. The training is carried out in three stages: first, the intermediate tool-calling model is trained on the seed data for 5 epochs with a learning rate of $5 \times 10^{-6}$; next, the embedding editor is trained on the full dataset for 10 epochs with a learning rate of $1 \times 10^{-4}$; finally, the LLM backbone and the embedding editor are jointly trained for an additional 2 epochs with a learning rate of $5 \times 10^{-6}$. The learning objective of the first stage is:
\begin{equation}
\begin{aligned}
\mathcal{L}^{\text{phase-1}}_{\mathrm{LM}}(\theta)
&= - \frac{1}{\sum_{i=1}^N |\mathcal{S}^{(i)}|}
    \sum_{i=1}^N \sum_{t=0}^{T^{(i)}-1} \\[6pt]
&\quad\;\;\sum_{\tau \in \mathcal{S}^{(i)}_t}
    \log P_\theta\!\left(x^{(i)}_\tau \,\middle|\, x^{(i)}_{<\tau},\, E^{(i)}_{v,\le t}\right),
\end{aligned}
\label{eqapp:lm_phase1_triple}
\end{equation}
which is mentioned in Section~\ref{section: model}. The learning objective of the second and third stages is:
\begin{equation}
\begin{aligned}
\mathcal{L}^{\text{phase-2}}_{\mathrm{LM}}(\theta)
&= - \frac{1}{\sum_{i=1}^N |\mathcal{S}^{(i)}|}
    \sum_{i=1}^N \sum_{t=0}^{T^{(i)}-1} \\[6pt]
&\quad\;\;\sum_{\tau \in \mathcal{S}^{(i)}_t}
\log P_\theta\!\left(x^{(i)}_\tau \,\middle|\, x^{(i)}_{<\tau},\, f_{\mathrm{editor}}\!\big(E^{(i)}_{v,\le t}\big)\right)
\end{aligned}
\label{eq:phase2-1}
\end{equation}

The main difference is that, in stages two and three, the VLM takes as input visual tokens produced by the editor instead of ground-truth visual context, and the LM objective is conditioned on the editor’s output.

\section{Discussions}
\label{sec:discussions}
\subsection{More Visualization Results on Public Benchmarks}
\label{app:roi}
In Section~\ref{embed editor}, we examined ``where the model is looking'' by visualizing difference maps on our in-house benchmark. Here, we extend this analysis with additional examples on public benchmarks in Fig.~\ref{fig:roi_public_fail} for a more comprehensive view. As shown in Fig.~\ref{fig:roipublic}, the difference maps generally align with the model’s textual edit intent. A notable case appears in the bottom-right example from MathVista, where the editor correctly attends to the faces of all three individuals in accordance with the instruction, despite the model being trained exclusively on code-rendered images that contain no such open-world scenarios. This result suggests that the model exhibits a certain degree of generalization, as it can attend to novel cases far beyond its training distribution. Then, we turn to failure cases in Fig.~\ref{fig:failroi}, where the model’s attention seems to deviate from the intent expressed in natural language instructions. These examples might demonstrate the current limitations of the editor and the remaining challenges in faithful visual manipulation.

\begin{table}[t]
\centering
\caption{Effect of decoupled multi-stage training. }
\resizebox{\linewidth}{!}{
\begin{tabular}{lccccccc}
\toprule
 \textbf{Setting} & \textbf{MathVerse} & \textbf{Wemath} & \textbf{MathVista} & \textbf{MathVision} & \textbf{LogicVista} & \textbf{Indicator-500} & \textbf{Average} \\
\midrule
 Single stage training & 39.6 & 36.9 & 66.2 & 25.7 & 45.6 & 39.1 & 42.2\\
Decoupled training  & 43.2 & 37.1 & 69.1 & 32.3 & 48.1 & 40.5 & 45.1\\
\bottomrule
\end{tabular}}
\label{tab:stt}
\end{table}

\begin{table}[t]
\centering
\caption{Ablation study: Comparison of Pure-text CoT versus Interleaved CoT strategies. The interleaved format consistently outperforms the text-only baseline, particularly on geometry-intensive benchmarks.}
\resizebox{\linewidth}{!}{
\begin{tabular}{lcccccc}
\toprule
 \textbf{Method} & \textbf{MathVerse} & \textbf{Wemath} & \textbf{MathVista} & \textbf{MathVision} & \textbf{LogicVista}  & \textbf{Average} \\
\midrule
 Pure Text CoT & 40.2 & 34.5 & 69.0 & 30.2 & \textbf{48.1}  & 44.4\\
 Interleaved CoT (Ours) & \textbf{43.2} & \textbf{37.1} & \textbf{69.1} & \textbf{32.3} & \textbf{48.1}  & \textbf{45.1}\\
\bottomrule
\end{tabular}}
\label{tab:puretext}
\end{table}

\begin{figure}[t]
    \centering
    \includegraphics[width=1\linewidth]{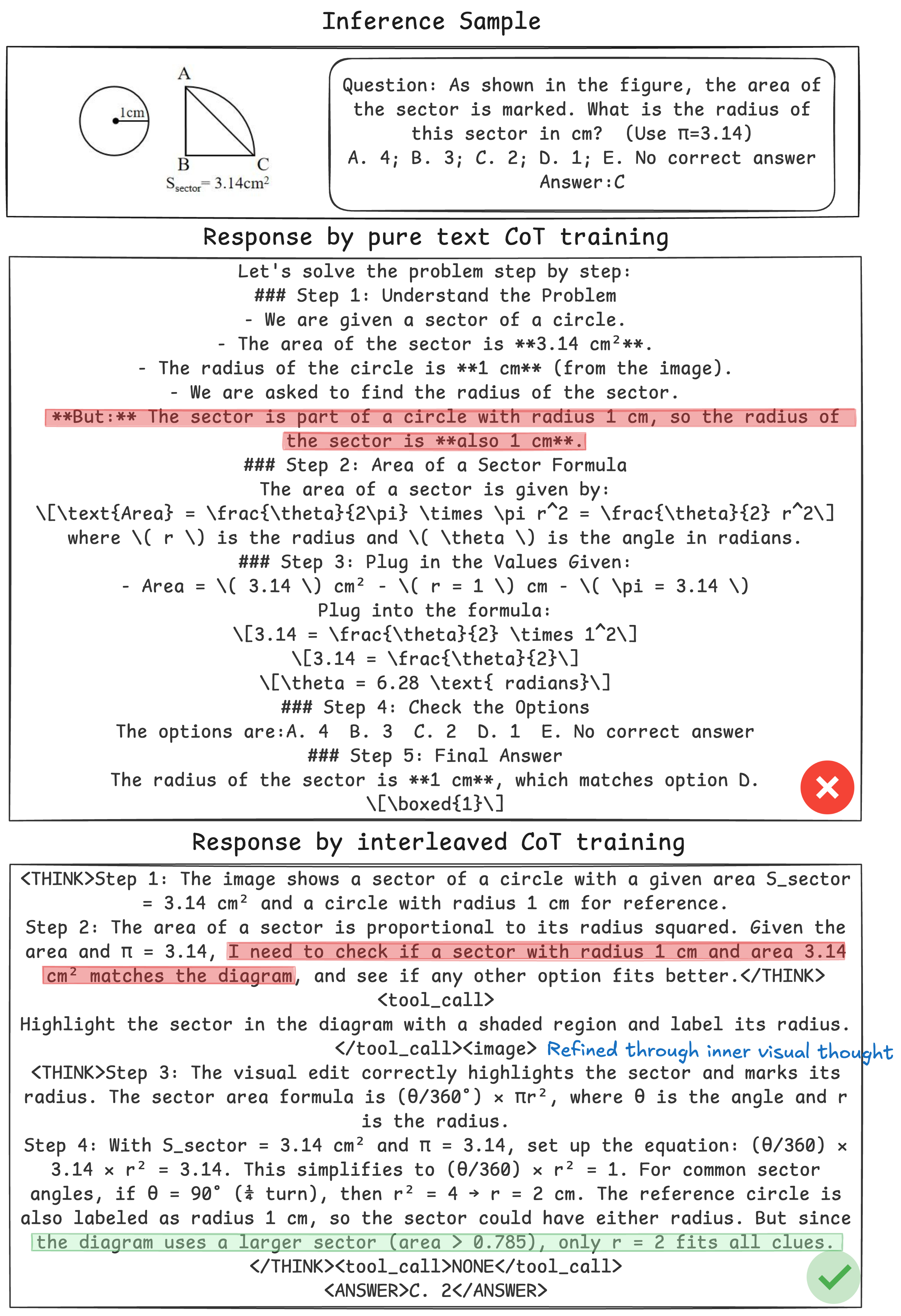}
    \caption{Inference sample and response from pure text CoT training and interleaved CoT training. The interleaved CoT refines the reasoning by correcting initial textual guesses through inner visual thought and tool-driven verification.}
    \label{fig:d_case1}
\end{figure}

\begin{figure}[t]
    \centering
    \includegraphics[width=1\linewidth]{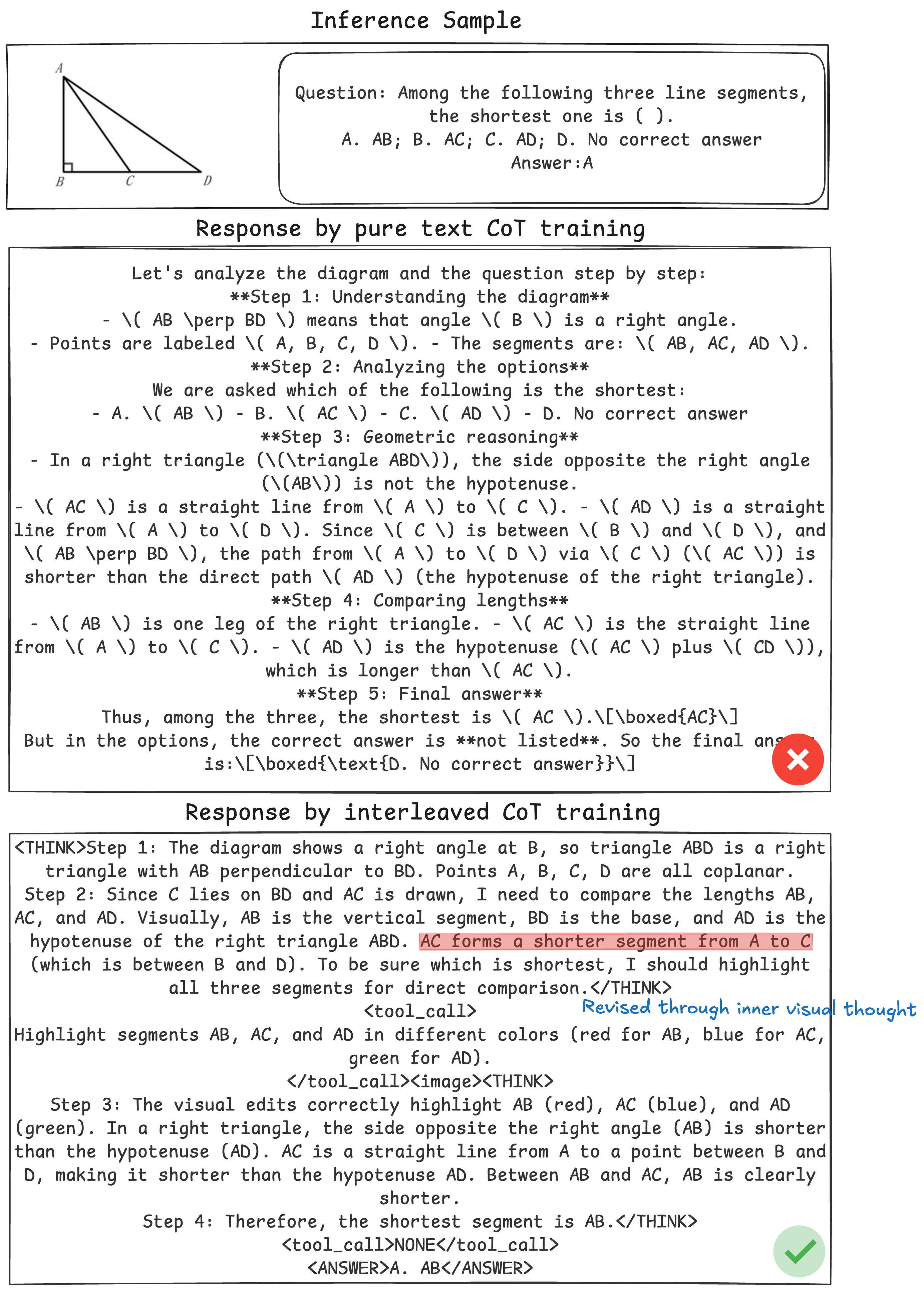}
    \caption{Inference sample and response from pure text CoT training and interleaved CoT training.}
    \label{fig:d_case2}
\end{figure}

\begin{table}[t]
\centering
\caption{
Accuracy comparison of different interleaved CoT strategies (Revise, Refine, Explore) against the pure-text baseline on the subset of 253 instances involving explicit tool usage.
}
\resizebox{1\linewidth}{!}{
\begin{tabular}{l l r r r}
\toprule
Strategy        & Model             & \#Correct & \#Total & Accuracy (\%) \\
\midrule
Revise          & Interleaved CoT   & 5         & 5       & 100.00 \\
Refine          & Interleaved CoT   & 50        & 126     & 39.68 \\
Explore         & Interleaved CoT   & 54        & 122     & 44.26 \\
\midrule
Overall         & Interleaved CoT   & 109       & 253     & 43.08 \\
Overall         & Pure text CoT     & 97        & 253     & 38.34 \\
\bottomrule
\end{tabular}
}

\label{tab:interleaved-strategies}
\vspace{-3mm}
\end{table}

\subsection{Ablation: Interleaved vs. Pure-text CoT}
We verify the necessity of visual-augmented reasoning by comparing our Interleaved CoT against a Pure-text CoT baseline. To ensure a fair comparison, both settings utilize the identical \emph{Solver} model and training data (DeepSketcher 31k), differing only in that the baseline is text-only CoT traces without access to visual tool invocation.

The results (Tab.~\ref{tab:puretext}) demonstrate that the interleaved format consistently outperforms the text-only approach (45.1\% vs. 44.4\% on average). Notably, the gains are substantial on geometry-heavy benchmarks like MathVerse (+3.0\%) and WeMath (+2.6\%), validating our hypothesis that incorporating visual intermediates significantly enhances the model's capability to handle spatially complex problems, where pure-text reasoning is prone to error. We present representative cases from WeMath in Fig.~\ref{fig:d_case1} and Fig.~\ref{fig:d_case2} to further illustrate the effectiveness of DeepSketcher. 

\begin{figure}
    \centering
    \includegraphics[width=0.8\linewidth]{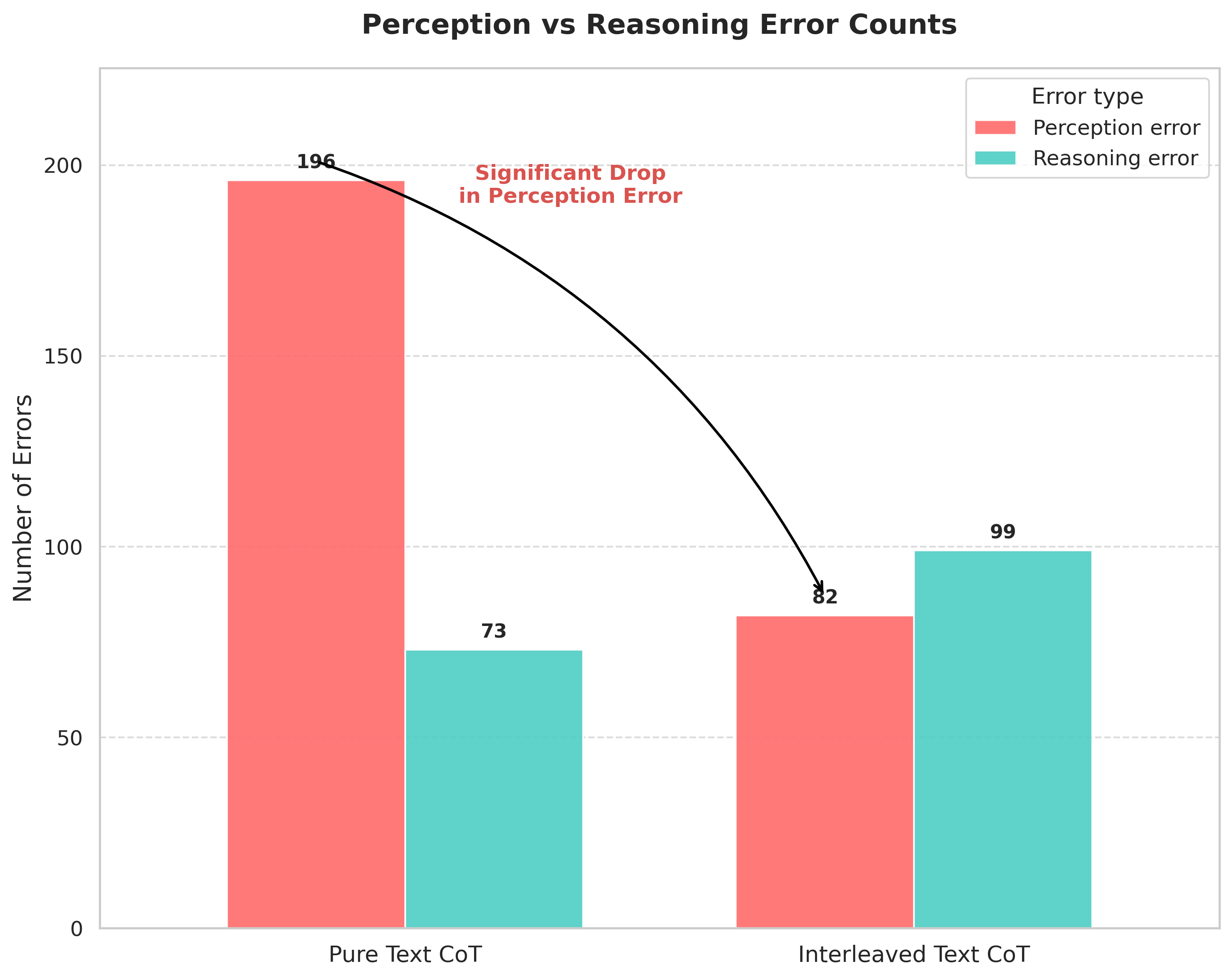}
    \caption{Error analysis on responses from pure text CoT training and interleaved CoT training.}
    \label{fig:err}
\end{figure}

To investigate the root causes of failure, we sample incorrect responses from the WeMath benchmark and conduct an error analysis. We categorized failures into two distinct groups: \textbf{Perception Errors}, where the model misidentifies visual elements or hallucinates non-existent details; and \textbf{Reasoning Errors}, where the model makes logical or calculation mistakes despite accurate visual grounding. As shown in Fig.~\ref{fig:err}, our analysis reveals that, compared to pure text CoT training, interleaved CoT training significantly reduces the prevalence of perception errors, demonstrating its effectiveness in grounding reasoning on valid visual evidence.

To deeper understand how the model employs visual tools and how the resulting visual feedback influences its logical flow and subsequent reasoning, we conduct a qualitative analysis on a sampled subset of responses from the MathVerse, WeMath, and MathVista benchmarks where the embedding editor is explicitly leveraged. We categorize the intended strategic purpose of these tool usages based on their impact on the reasoning trajectory: \textbf{Revise}: Tool usage leads to overturning previous assumptions or directions. \textbf{Refine}: Tool usage does not alter the overall direction but makes the subsequent reasoning more specific or rigorous. \textbf{Explore}: Tool usage is employed specifically to gather information (e.g., reading diagrams, labeling, or measuring) rather than for verification or revision. 

Table~\ref{tab:interleaved-strategies} presents the distribution and performance of these strategies. On this specific subset, our proposed DeepSketcher achieves an overall accuracy of 43.08\%. For comparison, a pure text CoT baseline evaluated on this identical instances yielded an accuracy of 38.34\%, indicating that visual intervention notably surpasses pure-text reasoning in these complex scenarios. (Notably, the sampling was conducted independently of baseline results).

A breakdown by strategy reveals interesting insights into the model's behavior. The \textit{Explore} strategy—gathering visual evidence prior to reasoning—is the most frequent and highly effective, reaching 44.26\% accuracy, clearly demonstrating the value of grounding reasoning in verified visual details. The \textit{Refine} strategy also demonstrates a slight advantage over the pure-text baseline (39.68\% vs. 38.34\%). Finally, although the \textit{Revise} strategy is relatively rare in this dataset (only 5 cases), it is decisive when employed, achieving a 100\% accuracy rate by successfully correcting initial errors.

\subsection{Effect of Decoupled Multi-Stage Training} 
To assess the effect of our multi-stage training strategy, we compare checkpoints with and without explicitly decoupling the training of LLM and the embedding editor. As shown in Tab.~\ref{tab:stt}, the three-stage training pipeline—which first pretrains the LLM’s reasoning ability, then introduces the embedding editor in a separate adaptation stage, and finally performs joint refinement—consistently outperforms the single-stage alternative.
This result suggests that decoupling the LLM from the editor during training is essential: it allows the base model to acquire robust reasoning skills before being exposed to the more complex task of interleaving reasoning with visual manipulation.

By contrast, directly training the entire system end-to-end in a single stage leads to weaker overall performance, likely because the model must simultaneously learn high-level reasoning and low-level embedding modification, increasing optimization difficulty and reducing stability.

\subsection{Analysis of MathVerse Results}
On the MathVerse (Vision-only) benchmark, our model achieves an accuracy of 43.2, outperforming both the baseline Qwen2.5-VL-7B and several tool-calling VLMs. Notably, Bagel-Zebra-CoT-7B attains an exceptionally high score on this benchmark, substantially surpassing our model and ranking near the top of the MathVerse Vision-only leaderboard, comparable to GPT-4.1. This strong performance can be partially explained by the fact that Bagel-Zebra-CoT-7B is post-trained on Bagel-7B~\citep{deng2025emerging}, whose base model already achieves a notably high score (45--50 according to our implementations) on the MathVerse benchmark. Therefore, the results of Bagel-Zebra-CoT-7B are in part a reflection of the capability of its foundation model. Despite this, our method consistently achieves a 2.4-point improvement in accuracy over five widely used benchmarks, further validating the effectiveness of our approach.

\section{Limitations and future work}
\label{sec:limitations}
The proposed DeepSketcher suite provides a complementary perspective to the ``thinking with images'' paradigm by curating a dataset constructed entirely from code—ensuring accuracy and avoiding the grounding and image-generation noise—and by designing a self-contained model that circumvents reliance on external APIs. Nevertheless, this solution comes with several inherent limitations.

First, the dataset is generated exclusively from code, which may limit the approach’s applicability to broader, open-world domains. Moreover, since all questions are automatically generated, there is little fine-grained control over aspects such as difficulty, style, or even the correctness of model-provided answers during reasoning. This lack of precision in data quality raises the risk of ``rubbish in, rubbish out,'' making it crucial to design comprehensive filtering pipelines to ensure the model learns from high-quality content. In this work, we introduced multiple filtering mechanisms and an \texttt{img2code} framework to mitigate these effects, but future efforts should focus on expanding data collection to more diverse and open-world domains while improving the quality of generated content.

Second, although our model design removes the dependence on external tools and the need to re-encode images repeatedly, it also diverges from unified generative understanding models in that its ``visual thoughts'' reside purely in the embedding space. As a result, the actual intermediate visual content remains inaccessible, limiting our ability to fully interpret and analyze the model’s behavior. Future work should explore more expressive ways to represent and manipulate visual information, thereby enhancing transparency and interpretability in reasoning within this paradigm.

\clearpage

